\documentclass[11pt]{article}
\usepackage{acl2014}
\usepackage{times}
\usepackage{latexsym}
\usepackage{url}
\usepackage{color}
\usepackage{multirow}

\usepackage{amsmath} 
\usepackage{amssymb}

\usepackage{etoolbox}
\usepackage{graphicx}
 
 \usepackage{enumitem}
  {\begin{itemize}[topsep=0pt, partopsep=0pt] \footnotesize%
    \setlength{\itemsep}{0pt}%
    \setlength{\parskip}{0pt}%
    }%
  {\end{itemize}}
  
  \usepackage{color}
  {\begin{enumerate}≈ \footnotesize%
    \setlength{\itemsep}{0pt}%
    \setlength{\parskip}{0pt}%
    }%
  {\end{enumerate}}
  
  \usepackage{CJKutf8}

\title{Voting for Deceptive Opinion Spam Detection}
\author{Tao Wang and Hua Zhu  \\
{\normalsize Computer Science Department, Nanjing University}\\
  {\normalsize Wuhan, Hubei, P.R. China, 430072}\\
    {\normalsize pwangtao@gmail.com, huaz@whu.edu.cn}
  }

\begin{document}
\maketitle
\begin{abstract}
Consumers' purchase decisions are increasingly influenced
by user-generated online reviews. Accordingly, there has
been growing concern about the potential for posting deceptive opinion spam
fictitious reviews that have been deliberately written to sound authentic, to deceive the readers.
Existing approaches mainly focus on developing automatic supervised learning based methods to help users identify deceptive opinion spams.  
 this work, we used the LSI and
Sprinkled LSI technique to reduce the dimension for deception detection. We make our contribution
to demonstrate what LSI is capturing in latent semantic space and reveal
how deceptive opinions can be recognized automatically from truthful opinions.
Finally, we proposed a voting scheme which integrates different approaches to
further improve the classification performance.
\end{abstract}

\section{Introduction}
Consumers increasingly rely on user-generated online reviews when making purchase decision \cite{Cone:11,Ipsos:12}. Unfortunately, the ease of posting content to the Web, potentially anonymously, create opportunities and incentives for unscrupulous businesses to post deceptive opinion spam fraudulent or fictitious reviews that are deliberately written to sound authentic, in order to deceive the reader \cite{Ott:11}.  For example, a hotel manager may post fake positive reviews to promote their own hotel, or fake negative reviews to demote a competitor`s hotel. Accordingly, 
there appears to be widespread and growing concern
among both businesses and the public \cite{Meyer:09,Miller:09,Streitfeld:12,Topping:10}. 
The followings are two reviews, one is deceptive and the other one is truthful.

\begin{enumerate}
\item {My husband and I arrived for a 3 night stay. for our 10th wedding anniversary. We had booked an Executive Guest room, upon ar- rival we were informed that they would be upgrading us to a beautiful Junior Suite. This was just a wonderful unexpected plus to our beautifully planned weekend. The front desk manager was professional and made us feel warmly welcomed. The Chicago Affinia was just a gorgeous hotel, friendly staff, lovely food and great atmosphere. Not the men- tion the feather pillows and bedding that was just fantastic. Also we were allowed to bring out beloved Shi-Tzu and he experienced the Jet Set Pets stay. The grooming was perfect, the daycare service we felt completely comfortable with. This was a beautiful weekend, thank you Affinia Hotels! We would visit this hotel again! }
\item {\em As others have said, once all the construction works are completed I suspect that the prices this hotel can (legitimately) charge will put it out of our price range. Which will be a pity because it is an excellent hotel, the location couldn’t be better. The room was very spacious, with separate sitting study areas a nice bathroom. Only 3 minor points: down- stairs they were serving a free basic break- fast (coffee and pastries), but we only knew of it on our last morning nobody had mentioned this; the cost of internet is a bit dear especially when lots of motels now offer it free; there was only powered milk in the rooms, which wasn’t that nice. But none of these really spolied a really enjoyable stay.. }
\end{enumerate}

As can be seen, as the distinction is too subtle, it is hard to manually tell truthful reviews from deceptive one (the first one is deceptive). 
Existing approaches for spam detection usually focus on developing supervised learning-based algorithms to help users identify deceptive opinion spam \cite{jindal2008opinion,jindal2010finding,li2011learning,lim2010detecting,Ott:11,wang2011review,wu2010distortion}. 
These supervised approaches suffer one main disadvantage: they are highly dependent upon high-quality gold-standard labelled data. One option for
producing gold-standard labels, for example, would be to
rely on the judgements of human annotators \cite{jindal2010finding,mukherjee2012spotting}. Recent studies, however, shows that unlike other kinds of spam, such as Web \cite{castillo2006reference,martinez2009web} and e-mail spam \cite{chirita2005mailrank}, deceptive opinion spam is neither easily ignored nor easily identified by human readers \cite{Ott:11}.
This is especially the case when considering the overtrusting nature of most human judges, a
phenomenon referred to in the psychological deception literature as a truth bias \cite{vrij2008detecting}. 

Due to the difficulty in manually labeling deceptive opinion, Ott et al. (2011) \cite{Ott:11} solicit deceptive reviews from workers on Amazon Mechanical Turk, and built a dataset containing 400 deceptive and 400 truthful reviews, which they use to train and evaluate supervised SVM classifiers\footnote{Truthful opinions are selected from 5-star reviews from TripAdvisor. http://www.tripadvisor.com/}. 
According to their findings, truthful hotel opinions are more specific about spatial configurations(e.g. small, bathroom, location), which can be easily explained by the fact that the Turkers have never been to that hotels. 

Our work in this paper started from 
the dataset, 400 truthful and 400 deceptive reviews, from \cite{Ott:11,li2013topicspam}, a large-scale, publicly available dataset for deceptive opinion spam research. Then
we use three ways to pre-processing dataset. (a) Use N-gram language model to build a probabilistic
language model (b) Use POS-tag to delve the linguistic reason why a review would likely to be
considered fake (c) Use TF-IDF to reflect how important a word is in given text.
After pre-processing, the data has high dimensional features which is quite computational expensive
and easily raises the over fitting problem. Thus we are motivated to reduce dimension using Latent
Semantic Indexing (LSI), one of most popular indexing and retrieval method that uses singular
value decomposition (SVD) to map the documents into vector in the latent concept-space. LSI
extracts useful information in latent concepts through learning a low-dimensional representation of
this text, which helps us to know more detailed difference in fake and real reviews. We also use
supervised latent semantic analysis (Sprinkle) to overcome the shortcoming of LSI and results show
performance improves a lot.
To classify data, we use Support Vector Machine (SVM) and Naive Bayes, two popular and mature
classifier, on our dataset after pre-processing and dimension reduction. Although these classifiers
are quite powerful, it may still misclassify correct points. So we use a voting scheme or a weighted
combination of the multiple hypothesis to achieve final conclusion.
The remainder of this paper is organized as follows: Section 2 talks about related work. Section
3 presents details of our fake reviews detection framework, including three approaches to preprocessing
data, two ways of dimension reduction, and two classification method. In section 4 shows
the experimental results and discussion of performance; Finally, Section 5 concludes our work.

\section{Related Work}
Jindal and Liu \shortcite{jindal2008opinion} first studied the deceptive opinion 
problem and trained models using features based on the review
text, reviewer, and product to identify duplicate opinions, i.e., opinions that
appear more than once in the corpus with similar contexts.
Wu et al. (2010) propose an alternative strategy
to detect deceptive opinion spam in the absence of a gold standard.
Yoo and Gretzel \shortcite{yoo2009comparison} gathered 40 truthful and 42
deceptive hotel reviews and manually compare the linguistic differences between them. 
Ott et al. created a gold-standard collection by employing Turkers to write fake reviews, 
and follow-up research was based on their data 
\cite{ott2012estimating,Ott-EtAl:2013:NAACL-HLT,li2013identifying,feng2013detecting,litowards}. 
For example, Song et al. \shortcite{feng2012syntactic} looked into syntactic features from Context Free Grammar parse trees
to improve the classifier performance. A step further, Feng and Hirst \shortcite{feng2013detecting} 
make use of degree of {\it compatibility} between the personal experiment and a collection of reference reviews
about the same product rather than simple textual features.

In addition to exploring text or linguistic features in deception, some existing work 
looks into customers' behavior to identify deception \cite{mukherjee2013spotting}.
For example, Mukherjee et al. 
\shortcite{mukherjee2011detecting,mukherjee2012spotting} 
delved into group behavior to identify 
group of reviewers who work collaboratively to write fake reviews.

\section{Data Processing}
Before building n-gram model, we tokenized the text of each review and used porter stemming
algorithms to eliminate the influence of different forms of one word. An n-gram model is a type of
2
probabilistic language model for predicting the next item in such a sequence in the form of n-1 order
Markov model. We built unigram and bigram model in our experiment.

\paragraph{tf-idf}
TF-IDF, short for term frequencyinverse document frequency, is a numerical statistic that is intended
to reflect how important a word is to a document in a collection or corpus. The tf-idf value increases
proportionally to the number of times a word appears in the document, but is offset by the frequency
of the word in the corpus, which helps to control for the fact that some words are generally more
common than others. For example, stop words – common words that appears every where but do
not have much meaning, such as the, a, and that, are of little significance. So tf-idf will assign small
weight to these words.

The formulae of tf-idf is:
$$w=tf\cdot idf_{i,j}=tf\cdot \log\frac{N}{df_{i,j}}$$

\paragraph{POS}
Part-of-speech tagging, also called grammatical tagging or word-category disambiguation, is the
process of marking up a word in a text as corresponding to a particular part of speech. Our project
use POS-tag to delve the linguistic reason why a review would likely to be considered fake. We use
the Stanford Parser to obtain the relative POS frequencies as feature vector to separate data.

\section{Model}
The power of dimension reduction lies in the fact that it can ease later processing, improve computational
performance improvement, filter useless noise and recover underlying causes. In our project,
we are using LSI for text classification. LSI is based upon the assumption that there is an underlying
semantic structure in textual data, and that the relationship between terms and documents
can be re-described in this semantic structure form. Textual documents are represented as vectors in
a vector space. It is fundamentally based on SVD that breaks original relationship of the data into
linearly independent components, where the original term vectors are represented by left singular
vectors and document vectors by right singular vectors.
That is
\begin{equation}
U_{d\times l}s_{l\times l}V_{l\times d}
\end{equation}
The column of $U_{d\times l}$ defines the lower dimensional coordinate system.
$S_l$ is a diagonal
matrix with the l largest singular values in non-increasing order along its diagonal. The l columns are the new coordinates of each document after dimensionality reduction. 

One major key of LSI is that it overcomes two of the most problematic constraints synonymy and
polysemy. It is very tolerant of noise (i.e, misspelled words, typographical errors, etc.) and has
been proven to capture key relationship information, including causal, goal-oriented, and taxonomic
information.Another usage of LSI is in data visualization. Let l=2 or 3. Each xi is approximated by
a l-dimensional vector now suitable for plotting in 2D or 3D spaces.

LSI has limitations in classification because it is an unsupervised method which doesn’t
take class information into account. So we decide to use the process Sprinkling in which LSI is
performed on a term-document matrix augmented with sprinkled terms, namely class labels.

\subsection{Classification}
\subsection{Naive Bayes} 
Naive Bayes classifiers are among the most successful known algorithms for learning
to classify text documents. As we know, text classifiers often don’t use kind of deep representation
about language: often a document is represented as a bag of words. This is a very simple representation
of document : it only knows which words are included in the document and throws away
the word order. NB classifier just relies on this representation. For a document D and a class C. In
our project, we only have two classes: fake or real. We classify d as the class which has the highest
posterior probability P(C|D), which can be re-expressed using Bayes Theorem:
\begin{equation}
p(c|d)=\frac{p(d|c)p(c)}{p(d)}
\end{equation}

\begin{equation}
\begin{aligned}
&C_{MAP}=argmax_c P(c|d)\\
&=argmax_{c}\frac{P(d|c)P(c)}{p(d)}\\
&=argmax_{c}P(d|c)P(c)
\end{aligned}
\end{equation}

\subsection{Support Vector Machine}
 Support Vector Machines(SVM) \cite{joachims1999making} are supervised learning algorithms that
used to recognize patterns and classify data. Given a set of binary training data, an SVM training
algorithm builds a model to calculate a hyper-plane that separate data into one category or the other.
The goal of SVM is to find a hyper-plane that clearly classify data and maximize the distance of the
support vectors to hyper-plane at the same time .
\begin{equation}
\frac{1}{2}||w||^2-\sum_{i}\alpha_i [y_i(w\cdot x_i-b)-1]
\end{equation}

We choose SVM as our classifier since in the area of text classification, SVM performs more robust
than other techniques. When implementing SVM, we first pre-process each original document
to a vector of features. Such text feature vectors often have high dimension, however, SVM
supports quite well. SVMs use overfitting protection, so the performance does not affect much when
the number of features increases. Joachims shows most text categorization problems are linearly
separable, so we use linear SVMs, which is also faster to learn, to solve our problem. In our project,
we use SVMlight to train our linear SVM models on datasets that processed by UNIGRAMS,
BIGRAMS, POS-Tag, and TF-IDF. We also train dataset after dimension reduction. Results show
that SVMs perform well on classifying our data.

\subsection{Voting}
Voting Scheme, also called weighted combination of the multiple hypothesis, is used to improve
performance when having different approaches to a certain problem.
Although some classifier is quite powerful, it may still misclassify correct points. On the other
hand, we desire simple models that free users from troublesome algorithm design. The uncertainty
of dimension selection and goodness evaluation makes classification difficult. This method is more
robust to difficult classified text and achieves better performance comparing to separate models.

\section{Experimental Results}
To better understand the models learned by these automated approaches, we decide to demonstrate
what SVD in LSI analysis is capturing. We find the group of words that have the most significant
influence in latent concepts. we can see, the second concept contributes most for
separating the fake reviews. We list in Table with the highest scores that loads
on the positive polar (truthful) and negative polar (fake) of second concept.
\begin{table}[!ht]
\centering
\begin{tabular}{ll}\hline
deceptive&truthful\\\hline
hotel&room\\
my&)\\
chicago&(\\
will&but\\
room&$\$$\\
very&bathroom\\
visit&location\\
husband&night\\
city&walk\\
experience&park\\\hline\hline
\end{tabular}
\caption{Top words in different topics from TopicSpam}
\end{table}

We
are using Sprinkled LSI and LSI to reduce the term-document matrix in different dimensions from
50 to 700. We use SVM to classify the data. The results show that with the dimension increasing,
the training accuracy of two methods is also increasing because we have higher features which give
us more information. Besides, the testing accuracy and precision is increasing when dimension is
from 0 to 500. When dimension continues to increase, the training accuracy increases while testing
accuracy goes down. It implies to us that there is an overfitting problem when the dimension of
the training data is high. Results in bold responds to best accuracy. We can see that when we use
Sprinkle + SVM and the dimension is around 500, the accuracy is 90$\%$ which is quite good.

In this paper, we combine 5 different algorithms above to automatically obtain the final classification
by voting for the test results. If over 3 methods vote for the same class, we will determine
the document to belong to that class. The five algorithms we choose as follows: Sprinkle+SVM
(dimension 500), Sprinkle+SVM(dimension 300), Unigram+SVM, TF.IDF+SVM, Unigram+NB.
The voting algorithm gets 0.95 accuracy.
\begin{table}
\centering
\begin{tabular}{|c|c|}\hline
SVM$_{unigram}$&0.90\\\hline
NB$_{unigram}$&0.863\\\hline
SVM$_{tf-idf}$&0.885\\\hline
SVM$_{sprinkle300}$&0.890\\\hline
SVM$_{sprinkle500}$&0.900\\\hline
Voting&0.95\\\hline
\end{tabular}
\caption{Performance comparisons of Voting Scheme vs separate algorithm}
\end{table}

\section{Conclusion}
In this work we use different methods to analyse and improve fake reviews classification result
based on Myle Ott and Yejin Choi work. It shows that the detection of fake reviews by using
automatically approaches is well beyond using human judges. We first build n-gram models and
POS model combined with SVM and Naive Bayes to detect the deceptive opinion. Similar to Myle
Ott and Yejin Chois result, n-gram based test categorization have the best performance suggesting
that key-word based approaches might be necessary in detecting fake review.
Besides, we used LSI to exert dimension reduction and some theoretical analysis. Contrast to Myle
Ott and Yejin Choi, our result shows that 2nd person pronouns, rather than 1st person form are
relatively tend to be used in deceptive review. The fake reviews also have the character of lacking of
the concrete noun and adjectives. Furthermore, we compare the experimental results of Sprinkled
LSI and LSI . We find Sprinkled LSI+ SVM outperforms LSI + SVM when dimension is not high.
Finally, we proposed a voting scheme which combines a few algorithms we used above to achieve
better classification performance.\bibliographystyle{acl}
\bibliography{acl2013}

\end{document}